\documentclass[cameraready]{Interspeech}
\newcommand{\ourmethod}{TurnGuide}

\title{TurnGuide: Enhancing Meaningful Full Duplex Spoken Interactions via Dynamic Turn-Level Text-Speech Interleaving}

\author[affiliation={1}]{Wenqian}{Cui}
\author[affiliation={2}]{Lei}{Zhu}
\author[affiliation={2}, correspondingauthor]{Xiao-Hui}{Li}
\author[affiliation={1}]{Zhihan}{Guo}
\author[affiliation={2}]{Haoli}{Bai}
\author[affiliation={2}]{Lu}{Hou}
\author[affiliation={1}, correspondingauthor]{Irwin}{King}

\address{
    $^1$ The Chinese University of Hong Kong \\
    $^2$ Huawei Technologies 
}

\email{wenqian.cui@link.cuhk.edu.hk}


\usepackage{comment}

\usepackage{enumitem}
\usepackage{xcolor}
\usepackage{pifont}
\usepackage{colortbl}
\usepackage{subcaption}
\usepackage{amsmath}
\usepackage{booktabs}
\usepackage{multirow}
\usepackage{listings}
\definecolor{beaublue}{rgb}{0.85, 0.9, 0.95}
\newtheorem{remark}{Remark}
\lstdefinestyle{mystyle}{
    backgroundcolor=\color{gray!20}, % Set the background color to a lighter gray
    basicstyle=\small\ttfamily,        % Set the font size to small and use typewriter font
    breaklines=true                     % Enable line breaking
}
\begin{document}

\maketitle

% the abstract here must exactly match the abstract entered into the paper submission system
\begin{abstract}
    % 1000 characters. ASCII characters only. No citations.
    % Full-Duplex Speech Language Models (FD-SLMs) are specialized foundation models designed to enable natural, real-time spoken interactions by modeling complex conversational turn-taking such as interruptions, backchannels, and overlapping speech. End-to-end (e2e) FD-SLMs leverage real-world double-channel conversational data to capture nuanced two-speaker dialogue patterns for human-like interactions, but their conversational abilities often degrade compared to pure-text conversation due to prolonged speech sequences and limited high-quality spoken dialogue data. Although interleaved text-speech generation could mitigate this degradation, integrating discrete text tokens into continuous double-channel audio streams could disrupt the precise time alignment required for fluid interaction. To address this, we propose \textbf{\mbox{\ourmethod{}}}, a novel text-speech interleaved generation approach for e2e FD-SLMs that dynamically segments assistant speech into dialogue turns and interleaves turn-level text and speech generation. This approach allows FD-SLMs to integrate the semantic intelligence of LLMs without compromising the natural acoustic flow. Extensive experiments show that \ourmethod{} not only significantly improves e2e FD-SLMs to produce semantically meaningful, coherent speech but also achieves state-of-the-art performance on various turn-taking events.
    Full-Duplex Speech Language Models (FD-SLMs) enable real-time spoken interaction by modeling turn-taking behaviors such as interruptions and overlapping speech. End-to-end (e2e) FD-SLMs learn from double-channel conversational audio to capture natural dialogue dynamics, but their conversational quality often degrades compared to text-only models due to long speech sequences and limited high-quality spoken data. Although interleaved text–speech generation may alleviate this issue, inserting discrete text tokens into continuous dual-channel audio can disrupt temporal alignment and interaction fluency. We propose \textbf{\mbox{\ourmethod{}}}, a turn-level text–speech interleaved framework that dynamically segments assistant speech and jointly generates text and speech. This enables semantic intelligence from LLMs while preserving natural acoustic flow. Experiments show that \ourmethod{} significantly improves semantic coherence and achieves state-of-the-art performance on diverse turn-taking events.\footnote{Code, models, and demo are available at \url{https://github.com/dreamtheater123/TurnGuide}.}
\end{abstract}

\section{Introduction}
% \begingroup
% \renewcommand\thefootnote{}\footnote{All the appendices are in the supplementary materials.}
% \addtocounter{footnote}{-1}
% \endgroup

\begin{figure*}[t]
    \centering
    \includegraphics[width=\textwidth]{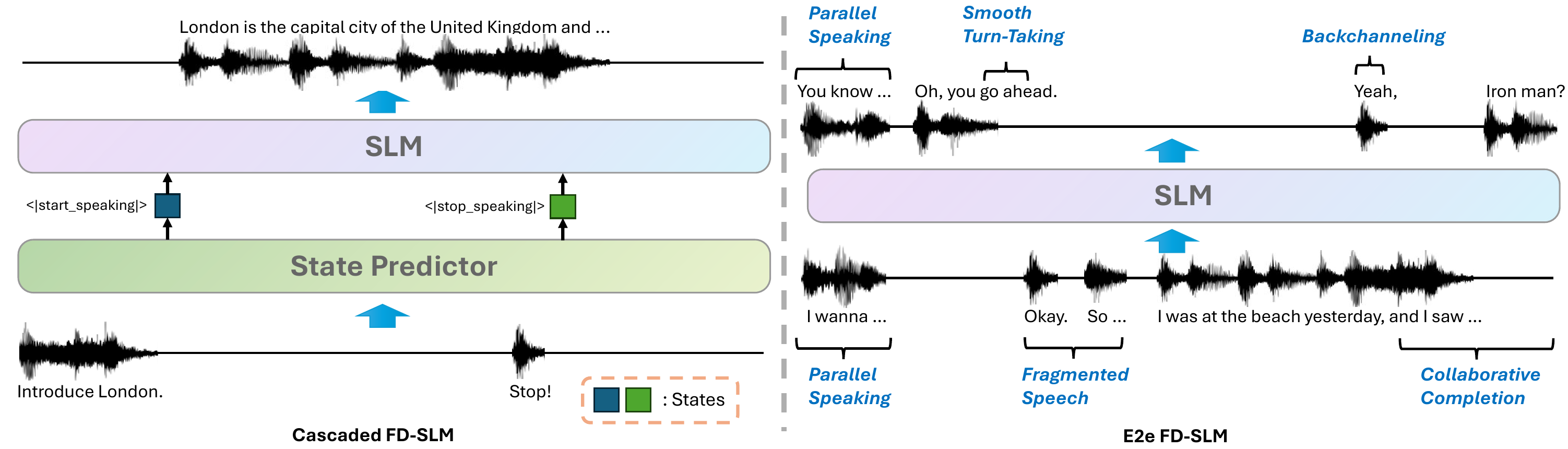}
    \caption{Illustration of the two types of FD-SLMs. Cascaded FD-SLMs utilize a state predictor to guide predefined behaviors, while e2e FD-SLMs capture the complex dynamics of real-world double-channel dialogue data.}
    % \vspace{-1.0\baselineskip} % Remove space after the table
    \label{fig:intro_figure}
\end{figure*}

\label{sec:intro}
Speech Language Models (SLMs) have attracted attention as foundation models for speech-based interactions with human users \cite{SLMsurvey,gpt4osystemcard}. Full-Duplex Speech Language Models (FD-SLMs), a specialized type of SLM, model user- and assistant-channel audio data simultaneously to support real-time communication with various natural turn-taking events such as interruptions \cite{minmo}, backchannels \cite{moshi}, and overlapping speech \cite{dgslm}, aiming to emulate human-like conversations \cite{dgslm,ntpp}. Existing FD-SLMs can be categorized as \textit{cascaded} and \textit{end-to-end (e2e)} systems. Cascaded FD-SLMs utilize a separate prediction module to classify spoken dialogues into predefined states, guiding SLM generation \cite{minmo,freezeomni}. While demonstrating efficacy in handling elementary turn-taking like interruptions, they are inherently bound by their dependence on explicit rule-based frameworks. In contrast, e2e FD-SLMs, which we adopt, leverage real-world conversational data to train models e2e, capturing nuanced dialogue patterns for more natural, human-like interactions \cite{ntpp,syncllm}. Figure \ref{fig:intro_figure} illustrates these two FD-SLM types.

While e2e FD-SLMs enable more natural conversational experiences, a key challenge is that their conversational abilities degrade compared to text-based Large Language Models (TLMs), due to prolonged speech sequences and the scarcity of high-quality speech training data \cite{glm4voice,miniomni}.
% While e2e FD-SLMs facilitate more natural conversational experiences, a significant challenge arises when training these models on spoken dialogue data. Their conversational abilities tend to degrade in comparison to text-based Large Language Models (TLMs) due to the nature of prolonged speech sequences and the scarcity of high-quality speech training data \cite{glm4voice,miniomni}. 
% One promising solution to this challenge is to transfer the text abilities to speech abilities by leveraging text-guided speech generation, where the model first generates the assistant's text response based on the input context, followed by generating the corresponding speech response. However, applying text guidance in e2e FD-SLMs poses significant challenges, as these models require precise time alignment between the user and assistant audio streams to effectively capture complex conversational patterns and provide timely, meaningful responses. Integrating text elements into the audio streams can easily disrupt this alignment, ultimately compromising the responsiveness and coherence of the generated speech.
One promising solution is to transfer text abilities to speech abilities via text-speech interleaved generation, where the model first generates the assistant’s text response and uses it to guide subsequent speech generation. However, applying text guidance in e2e FD-SLMs poses significant challenges, as these models require precise time alignment between user and assistant audio streams to capture complex conversational patterns and provide timely, meaningful responses. Interleaving text tokens into the audio streams can easily disrupt this alignment, compromising the responsiveness and coherence of the generated speech.

% The challenges of integrating texts into assistant-channel audio data can be summarized into two main aspects. \textbf{1) Insertion Timing.} The text must be inserted precisely at the moment the assistant begins speaking. If the insertion is too early, the assistant may respond without accessing the full context of the user's speech, leading to misinterpretations or hallucinations. Conversely, if the insertion is too late, the assistant's speech may precede the inserted text, rendering the text guidance approach ineffective. \textbf{2) Insertion Length.} The length of the inserted text is critical. If the text is too lengthy, the assistant may continue speaking excessively and struggle to adapt its responses to subsequent user inputs. On the other hand, if the inserted text is too brief, the semantic content may become overly fragmented, diminishing the effectiveness of the text guidance approach \cite{roboego}.

The challenges of integrating text into assistant-channel audio data can be summarized in two aspects. \textbf{1) Insertion timing.} Text must be inserted precisely when the assistant begins speaking. If it is inserted too early, the assistant may respond without the full user context, causing hallucinations. If it is inserted too late, the assistant’s speech may precede the text, undermining the text guidance's effectiveness. \textbf{2) Insertion length.} The length of the inserted text is also critical. If it is too long, the assistant may overspeak and fail to adapt to subsequent user inputs; if too short, the semantic content becomes fragmented, weakening the effectiveness of text guidance \cite{roboego}.

To address these challenges, we propose a novel approach, \textbf{\ourmethod{}}, to seamlessly integrate text data into \textbf{real-world} double-channel spoken conversational data for the training of e2e FD-SLMs. Specifically, we dynamically segment the assistant's speech into \textbf{turns}, which represent the reasonable lengths of spoken content that convey the assistant's intended expression. Then, the assistant is trained to interleave the text-speech generation for each turn for proper guidance. By doing so, our method effectively addresses both insertion timing and insertion length challenges. Extensive experiments show that our approach significantly improves the conversational abilities of e2e FD-SLMs by not only producing \textbf{semantically meaningful and coherent speech} but also achieving state-of-the-art performance on various \textbf{turn-taking events}. To summarize our contributions:
% \vspace{-5pt}
\begin{itemize}[left=0pt]
    \item We propose \textbf{\ourmethod{}}, a novel text-speech interleaved generation approach for e2e FD-SLMs grounded in natural speech data.
    \item We introduce a dynamic turn segmentation framework and a text-guided dialogue modeling framework within \ourmethod{} to address the insertion timing and length in double-channel dialogues.
    \item We demonstrate through extensive experiments that \textbf{\ourmethod{}} substantially improves not only FD-SLMs' ability to produce semantically meaningful, coherent speech but also the behaviors on various turn-taking events.
\end{itemize}
% while preserving the model's ability to engage in timely and interactive conversations.
% while maintaining the natural conversational flow of the full duplex spoken dialogue.
% while preserving the model's ability to engage in interactive conversations with natural flow.

% About ensuring efficient text guidance, there are two points to claim. First, we ensure a minimum number of text tokens to insert, which requires a significantly smaller number of predicted tokens during inference than Moshi's approach. This significantly reduces the inference cost by x percent. Second, there is a latency upper bound.

\section{Related Works}
\subsection{Full Duplex Speech Language Models}
FD-SLMs process double-channel conversational data to interact with users seamlessly and can be categorized into \textit{cascaded} and \textit{end-to-end (e2e)} approaches. 

\subsubsection{Cascaded FD-SLMs}
Cascaded FD-SLMs utilize a separate prediction module to classify full-duplex dialogue into a finite number of predefined states, which are typically designed to facilitate transitions between listening and speaking to guide SLM behavior. Training cascaded FD-SLMs consists of two key components: a state predictor, which predicts the SLM's next state, and a backbone SLM, typically trained on synthetic conversational data. Therefore, cascaded FD-SLMs are constrained by their predefined behaviors, preventing them from authentically engaging in human-like spoken conversations.

The main focus of cascaded FD-SLMs is on defining states. For instance, LSLM \cite{lslm} uses states like ``interrupted'' and ``keep speaking'', where interruptions are triggered by specific keywords. Mini-Omni 2 \cite{miniomni} specifies ``start speaking'' and ``stop speaking'', halting generation upon hearing ``Stop Omni!''. Freeze-Omni \cite{freezeomni} bases states on whether the user has finished speaking, prompting the model to wait or interrupt. Minmo \cite{minmo} simplifies its states to whether the model should speak up or keep listening. These methods primarily rely on synthetic conversational data to manage state transitions, limiting their ability to foster natural, human-like spoken interactions.

\subsubsection{End-to-end FD-SLMs}
E2e FD-SLMs, on the other hand, employ a transformer-based foundation model architecture \cite{transformer,gpt4} to learn from real-world double-channel conversational audio data. This e2e approach enables SLMs to autonomously capture intricate conversational dynamics from authentic human interactions, enabling more natural conversations.

The main focus of e2e FD-SLMs is on effectively modeling double-channel audio data and capturing the interactions between these channels. dGSLM \cite{dgslm} introduces a Dialogue Language Model that employs distinct transformers to process each audio channel separately, using cross-attention mechanisms to model their interactions. SyncLLM \cite{syncllm} takes a different approach by jointly modeling the double-channel audio within a single LLM, interleaving the two audio sequences into a unified sequence. Moshi \cite{moshi} adopts an RQ-Transformer architecture \cite{rq-transformer}, designed to predict the assistant’s output based on the user’s input at each time step. NTPP \cite{ntpp} proposes a Next-Token-Pair Prediction training objective, enabling simultaneous prediction of audio tokens for both channels in each step.

\subsection{Text-Guided Speech Generation in SLMs}
SLMs enable e2e spoken interactions by generating output speech based on user speech input. However, studies have shown that SLMs exhibit significantly lower performance compared to TLMs counterparts \cite{voxeval,minmo,vitaaudio}. To mitigate this, researchers focus on integrating text to guide the speech generation process, aiming to enhance SLMs by transferring capabilities from TLMs.

Transferring abilities from TLMs to SLMs involves two key processes: representation alignment and text-speech interleaved generation. Representation alignment ensures the connection between text and speech representations within the SLM backbone, enabling knowledge transfer from text modality to speech modality. This alignment is commonly achieved through Automatic Speech Recognition (ASR) and Text-To-Speech (TTS) tasks during SLM pre-training \cite{glm4voice,freezeomni,minmo}. Some approaches also partially replace speech data with its corresponding text in the pre-training data \cite{spiritlm, glm4voice}. 
Subsequently, SLMs are trained to answer spoken queries by interleaving the generation of textual output with speech output, thereby effectively leveraging text information to enhance the quality of spoken responses \cite{glm4voice,moshi}.
% Following this, SLMs are trained to produce textual output before the final speech output, effectively leveraging text information for better spoken response quality.
% using speech interaction datasets to achieve seamless end-to-end capabilities for speech-based interactions, effectively enhancing their performance and functionality.

The exploration of text-guided speech generation for e2e FD-SLMs, particularly those trained on real-world full duplex speech data, is still in its early stages. Currently, the only notable study in this area is Moshi \cite{moshi}, which proposes adding text tokens corresponding to their respective speech sounds as they occur. However, this method fragments text semantics excessively and does not significantly improve SLM performance. In contrast, our approach aggregates the text semantics and inserts text by turns, achieving substantial enhancements in SLM performance. We acknowledge that some e2e FD-SLMs, like \cite{salm-duplex}, are trained on synthetic data. However, using synthetic data limits their ability to learn from genuine human dialogue patterns, making them outside the scope of our discussion.

\section{Methodology}
\begin{figure*}[t]
    \centering
    \includegraphics[width=0.99\textwidth]{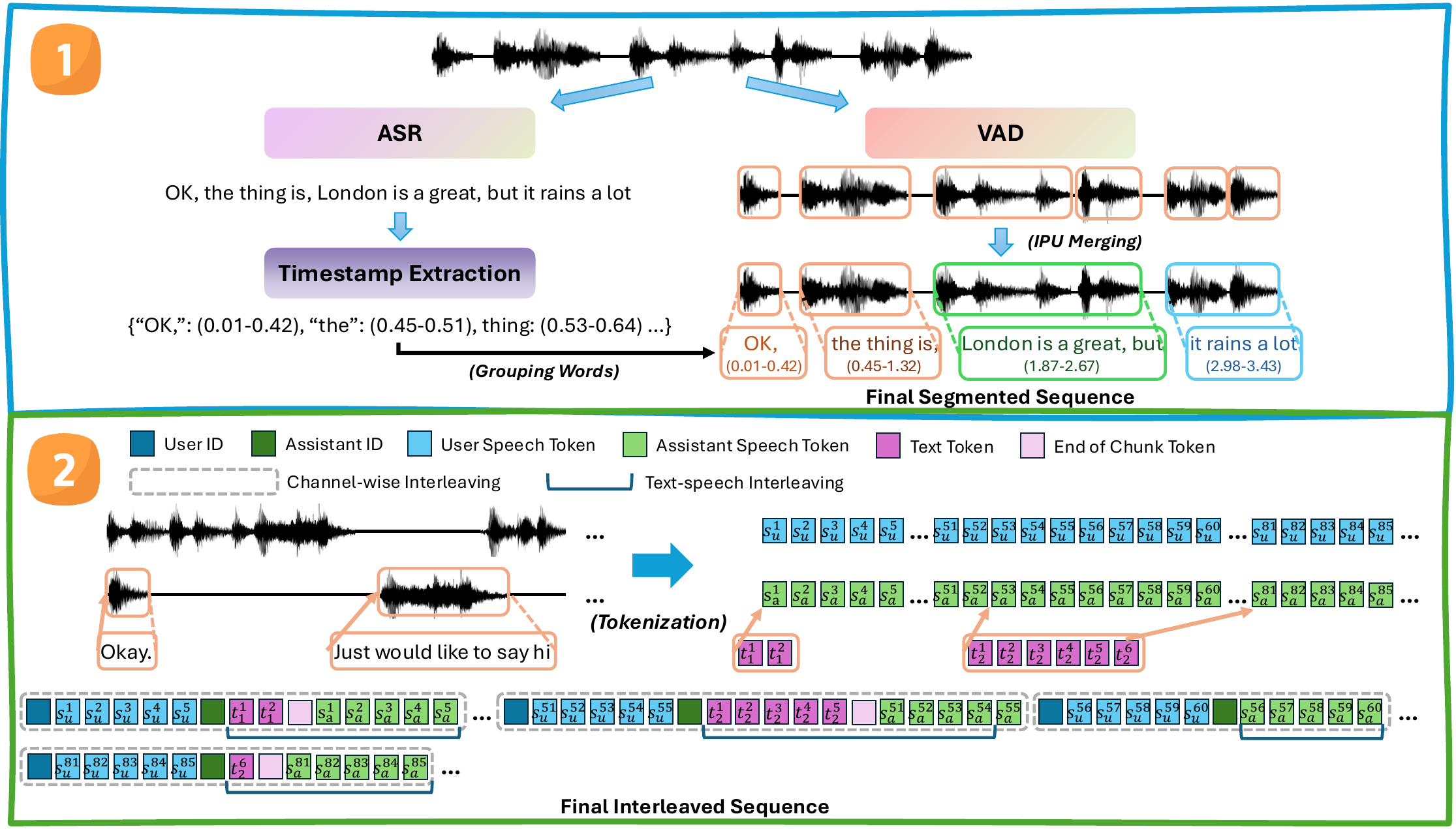}
    \caption{Illustration of the \ourmethod{} approach. The first part illustrates the multi-modal turn segmentation and alignment framework, which dynamically segments assistant speech into turns and aligns the text with speech turns. The second part shows the text-guided full duplex dialogue modeling framework with two interleaving strategies.}
    % \vspace{-1.0\baselineskip} % Remove space after the table
    \label{fig:method_overall}
\end{figure*}

We present the comprehensive methodology of \ourmethod{}, designed to enhance e2e FD-SLMs by utilizing text guidance on real-world speech data. This approach is divided into two key frameworks: one for dynamically segmenting spoken dialogue into turns and performing speech-text data alignment, and the other for modeling text-guided full-duplex dialogue.

\subsection{Dynamic Turn Segmentation and Alignment \mbox{Framework}}
\label{sec:turnsegmentation}
To enable text guidance in full-duplex speech interactions, it is essential to derive training data that effectively combines text and speech. As a first step, it is crucial to determine how text can be incorporated without disrupting the natural conversational flow. This requires the text to be introduced at an appropriate moment and kept within a reasonable length. We propose tackling these challenges by segmenting the speech into turns and aligning each part of the text with its corresponding speech.

As shown in the first part of Figure \ref{fig:method_overall}, the framework starts by applying Voice Activity Detection (VAD) to identify speech segments within the audio stream. Let $A$ represent the input audio sequence. VAD produces a list of speech segments:
\begin{equation}
    % \small
    V = \text{VAD}(A) = \{(\tau_1^{start}, \tau_1^{end}), (\tau_2^{start}, \tau_2^{end}), ...\},
\end{equation}
where each tuple $(\tau_i^{start}, \tau_i^{end})$ represents the start and end timestamps of the $i$-th detected speech segment. We use the \textit{pyannote} library \cite{pyannote} as the VAD module. Considering the VAD result is too fragmented, we merge nearby speech segments into Inter-Pausal Units (IPUs) \cite{dgslm}, which are defined by consecutive segments having pauses within a threshold $\tau_{IPU}=0.5$ seconds, i.e., 
\begin{equation}
    IPU_j = \bigcup_{i \in G_j} (\tau_i^{start}, \tau_i^{end}),
\end{equation}
where $G_j = \{i: \tau_{i+1}^{start} - \tau_i^{end} < \tau_{IPU}\}$ defines the group of consecutive segments to be merged into IPU $j$.

Enabling real-time text guidance requires the precise alignment of text and speech data. We use an Automatic Speech Recognition (ASR) module with word-level timestamp functionality to acquire the time-aligned transcript. While some speech dialogue datasets already provide time-aligned text data, our approach ensures adaptability to unlabeled dialogue datasets. The ASR process is formulated as
\begin{equation}
    W = \text{ASR}(A) = \{w_1, w_2, ...\},
\end{equation}
where $w_i$ represents the $i$-th word in the transcription. The word-level timestamp extraction process is formulated as
\begin{equation}
    % \small
    \Theta = \text{TE}(A) = \{(\delta_1^{start}, \delta_1^{end}), (\delta_2^{start}, \delta_2^{end}), ...\},
\end{equation}
where $(\delta_i^{start}, \delta_i^{end})$ represents the start and end timestamps of word $w_i$. In practice, we employ the Whisper medium model \cite{whisper} for both ASR and timestamp extraction tasks. To enhance the stability of the timestamp extraction process, we utilize the \textit{whisper-timestamped} package \cite{whisper-timestamped}.

% Finally, timestamp information is utilized to refine turn segmentation. Words are grouped into their respective IPUs based on $ \tau ^ { start } $, which guides the grouping process. To address potential misalignment issues in timestamping, a tolerance parameter $ \tau _ { tol } = 0.6 $ seconds is introduced, allowing slight adjustments to segmentation boundaries. This ensures that instances where the first word of an IPU erroneously aligns with the preceding IPU are corrected. The overall process is defined as follows:

Finally, we align and group each word into its associated IPU using $\tau^{start}$. Additionally, we incorporate a tolerance parameter $\tau_{tol}=0.6$ seconds to adjust the word segmentation boundary slightly. This adjustment accounts for instances where, in practice, the first word in an IPU might incorrectly align with the preceding IPU due to inconsistencies in timestamping. The process is defined as
\begin{equation}
    % \small
    W_j = \{w_i : \tau_{IPU_j}^{start} - \tau_{tol} \leq \delta_i^{start} \leq \tau_{IPU_{j+1}}^{start} - \tau_{tol}\},
\end{equation}
where $W_j$ represents words grouped into turn $j$, and $\tau_{IPU_j}^{start}$ and $\tau_{IPU_{j+1}}^{start}$ represent the start times of IPU $j$ and $j+1$. We denote the speech segment corresponding to IPU $j$ as $S_j$. The final output consists of aligned text-speech pairs $\{(W_{j}, S_{j})\}$ where each text segment is temporally aligned with its corresponding speech segment, serving as training data for text-guided full-duplex dialogue modeling.

% ... The spoken turn segmentation can be viewed as an information optimization problem.

% Our approach makes the assumption that humans plan the content to say for each turn, and they say the whole turn out no matter what the other speaker says.

% By adding text to speech, we imitate the human planning process in spoken conversations.

% determine whether to include an algorithm or not

\subsection{Text-Guided Full Duplex Dialogue Modeling \mbox{Framework}}
\label{sec:dialoguemodeling}
After deriving the aligned dialogue turns, the next step is to model double-channel audio data with text guidance. We use GLM-4-Voice \cite{glm4voice}\footnote{\url{https://huggingface.co/zai-org/glm-4-voice-9b}}, an off-the-shelf SLM, as the backbone for training on full-duplex dialogue datasets. GLM-4-Voice includes a speech tokenizer to convert audio into discrete tokens, a language model to process these tokens, and a vocoder to reconstruct speech. Its strong speech-text alignment makes it ideal for the backbone of text-guided speech interaction. As illustrated in the second part of Figure \ref{fig:method_overall}, our text-guided full duplex dialogue modeling framework centers around \textbf{two interleaving strategies}:

% Our text-guided full duplex dialogue modeling framework is illustrated in Figure \ref{fig:method_full_duplex_modeling}, which centers around two interleaving strategies:

\textbf{Channel-wise Interleaving.} Full-duplex dialogue modeling necessitates the simultaneous processing of two-channel audio to capture cross-channel interactions. A common technique to address this is by interleaving the two audio channels into a single sequence, either with token-level interleaving \cite{ntpp} or chunk-level interleaving \cite{syncllm,omniflatten}. We choose chunk-level over token-level interleaving because it demonstrates better performance in our experiments. We first consider channel-wise interleaving for speech tokens. To implement this technique, let $s_u$ and $s_a$ represent speech tokens from the user and assistant channels. Each channel’s speech tokens are divided into speech chunks with a chunk size of $\tau_{SC} = 5$. Given GLM-4-Voice’s frame rate of $fr = 12.5\text{Hz}$, each speech chunk corresponds to a 400-ms audio segment. The chunk-interleaving process alternates segments between the two channels, forming a sequence $SC = [SC_u^1, SC_a^1, SC_u^2, SC_a^2, \dots]$. Each speech chunk contains $\tau_{SC}$ speech tokens, with a unique speaker ID token inserted at the beginning to distinguish between the channels. Specifically, $SC_u^i = [UID, s_u^{i\times\tau_{SC}+1}, s_u^{i\times\tau_{SC}+2}, \dots, s_u^{i\times\tau_{SC}+\tau_{SC}}]$ and $SC_a^i = [AID, s_a^{i\times\tau_{SC}+1}, s_a^{i\times\tau_{SC}+2}, \dots, s_a^{i\times\tau_{SC}+\tau_{SC}}]$ represent the $i$-th user and assistant channel speech chunks, where $UID$ and $AID$ denote the user and assistant ID tokens.

\textbf{Text-speech Interleaving.} We apply text guidance on the assistant channel. Specifically, we interleave the text tokens with speech tokens for each turn, starting from the first speech chunk of the turn. 

Given the text in a turn $W_{j}$, we first split it into text chunks $TC_j = [TC_j^1, TC_j^2,...]$ with chunk size $\tau_{TC} = 5$, where $TC_j^k = [t_j^{k\times\tau_{TC}+1}, t_j^{k\times\tau_{TC}+2}, ..., t_j^{k\times\tau_{TC}+\tau_{TC}}]$ represents the $k$-th text chunk for turn $j$. Then, we append an ``end of chunk'' ($\langle EOC \rangle$) token to the end of the text chunks. This can help the model to learn the dynamic turn boundary information. Second, we calculate the corresponding speech chunk for text-speech interleaving for each text chunk $TC_j^k$ in $W_j$ as
\begin{equation}
    % \small
    k_j = \lfloor \frac{\tau_{j}^{k0}\times fr}{\tau_{SC}} \rfloor,
\end{equation}
where $\tau_{j}^{k0}$ represents the timestamp for the first text token of the $k$-th text chunk in turn $j$. The text-speech interleaving is performed at the $k_j$-th chunk of every turn $j$.

The final sequence is formed by chunks with the two interleaving strategies, $FC = [C_u^1, C_a^1, C_u^2, C_a^2, ...]$, where
% \begin{equation}
%      FC = [C_u^1, C_a^1, C_u^2, C_a^2, ...],
% \end{equation}
% where
% $C_u^i = SC_u^i$ and
% \begin{equation}
% \small
% C_u^i = SC_u^i \text{ and } C_a^i = \begin{cases}
% [TC_j^{k}, SC_a^i] & \text{if text chunk presents} \\
% SC_a^i & \text{otherwise}.
% \end{cases}
% \end{equation}
\[
% \scriptsize
C_u^i = SC_u^i \text{ and } C_a^i = \begin{cases}
[TC_j^{k}, SC_a^i] & \text{if text chunk presents} \\
SC_a^i & \text{otherwise}.
\end{cases}
\]
The decision to use a 5:5 ratio for text chunks and speech chunks ensures that text generation precedes speech generation, providing effective text guidance. In summary, our full duplex dialogue modeling framework leverages two interleaving strategies to enable both cross-channel interaction and text-guided speech modeling.

\begin{table}[t]
  \centering
  \caption{\label{tab:opensource}
    Summary of the open-source status of previous e2e FD-SLM papers.
    % up to two weeks before the submission deadline.
  }
  % \rowcolors{1}{white}{lightblue}
  % \small % or \footnotesize or \scriptsize
\scalebox{0.9}{
  \begin{tabular}{lcc} % Adjust the width as needed
    \toprule
    \textbf{SLMs}  & \textbf{Fisher-finetuned}  & \textbf{Open-source} \\
    \midrule
    % \rowcolor{lightblue} \multicolumn{4}{c}{\textbf{Direct Answer (String-matching)}} \\ % Single word centered across all columns
    dGSLM \cite{dgslm}      & \textcolor{green}{\ding{51}} & \textcolor{green}{\ding{51}} \\
    SyncLLM \cite{syncllm}      & \textcolor{green}{\ding{51}} & \textcolor{red}{\ding{55}} \\
    Moshi \cite{moshi}      & \textcolor{red}{\ding{55}} & \textcolor{green}{\ding{51}} \\
    OmniFlatten \cite{omniflatten}      & \textcolor{green}{\ding{51}} & \textcolor{red}{\ding{55}} \\
    NTPP \cite{ntpp}      & \textcolor{green}{\ding{51}} & \textcolor{red}{\ding{55}} \\
    Wu et al. \cite{FD-SLM-Align}      & \textcolor{red}{\ding{55}} & \textcolor{red}{\ding{55}} \\
    SALM-Duplex \cite{salm-duplex}      & \textcolor{red}{\ding{55}} & \textcolor{red}{\ding{55}} \\
    \bottomrule
  \end{tabular}
}
  % \vspace{-1.0\baselineskip} % Remove space after the table
\end{table}

% \begin{table}[t]
% \centering
% % \rowcolors{1}{white}{lightblue}
% % \small % or \footnotesize or \scriptsize
% \scalebox{0.74}{
% \begin{tabular}{lc|cccc} % Added vertical line after second column
% \toprule
% \multirow{2}{*}{\textbf{SLM}} & \multicolumn{1}{c|}{\textbf{Semantics}} & \multicolumn{4}{c}{\textbf{Turn-taking (Pearson Correlation)}} \\
% \cmidrule(l){2-2} \cmidrule(l){3-6}
%  & \textbf{GPT-score} & \textbf{Occ}  & \textbf{CumDur} & \textbf{AvgDur} & \textbf{Overall} \\
% \midrule
% % \rowcolor{lightblue} \multicolumn{4}{c}{\textbf{Direct Answer (String-matching)}} \\ % Single word centered across all columns
% dGSLM       & 1.32 & 0.31 & 0.24 & 0.13 & 0.23 \\
% STI        & 2.55 & \textbf{0.32} & 0.29 & 0.14 & 0.25 \\
% SCI       & \textbf{2.71} & 0.31 & \textbf{0.35} & \textbf{0.17} & \textbf{0.28} \\
% \bottomrule
% \end{tabular}
% }
% \caption{\label{tab:allbaseline}
% Comparison of both the semantic (GPT-score, measured on Fisher dataset) and turn-taking abilities (measured on Candor dataset) between our trained baselines and dGSLM. Occ, CumDur, and AvgDur represent occurrence, cumulative duration, and averaged duration, respectively.
% }
% \end{table}

\begin{table*}[t]
\centering
\caption{\label{tab:semanticeval}
Semantic evaluation results (0-10↑) of different FD-SLMs under different temperature settings by GPT-4o.
}
\scalebox{0.85}{
\begin{tabular}{lccccc|ccccc|ccccc|c} 
\toprule
\multirow{2}{*}{\textbf{SLM}} & \multicolumn{5}{c|}{\textbf{Unconditional}} & \multicolumn{5}{c|}{\textbf{Conditional - All}} & \multicolumn{5}{c|}{\textbf{Conditional - Assistant}} & \multirow{2}{*}{\textbf{Overall}} \\
\cmidrule(lr){2-6} \cmidrule(lr){7-11} \cmidrule(lr){12-16}
& \textbf{0.8} & \textbf{0.9} & \textbf{1.0} & \textbf{1.3} & \textbf{AVG} & \textbf{0.8} & \textbf{0.9} & \textbf{1.0} & \textbf{1.3} & \textbf{AVG} & \textbf{0.8} & \textbf{0.9} & \textbf{1.0} & \textbf{1.3} & \textbf{AVG} & \\
\midrule
dGSLM & 4.70 & 5.00 & 4.94 & 4.03 & 4.67 & - & - & - & - & - & - & - & - & - & - & - \\
STI & 4.06 & 4.94 & 6.00 & 7.79 & 5.70 & 3.58 & 4.36 & 4.94 & 6.18 & 4.76 & 2.76 & 3.36 & 3.76 & 5.24 & 3.79 & 4.76 \\
SCI & 5.48 & 6.58 & 7.52 & 8.21 & 6.94 & 5.15 & 5.67 & 6.00 & 6.55 & 5.85 & 4.09 & 4.61 & 5.06 & 6.06 & 4.94 & 5.91 \\
Moshi & - & - & - & - & - & 6.70 & 6.39 & 5.94 & 4.21 & 5.82 & 6.39 & 6.42 & 6.30 & 4.24 & 5.85 & - \\
Moshi TS & 5.21 & 6.24 & 7.30 & 8.30 & 6.76 & 4.42 & 4.94 & 5.52 & 6.45 & 5.33 & 3.73 & 4.18 & 4.76 & 5.94 & 4.67 & 5.58 \\
\ourmethod{} & 8.12 & 8.76 & 8.88 & 8.61 & 8.61 & 6.24 & 6.58 & 7.09 & 7.15 & 6.76 & 6.12 & 6.42 & 6.58 & 6.79 & 6.48 & 7.27 \\
\ourmethod{} L2:1 & 8.70 & \textbf{9.27} & \textbf{9.33} & \textbf{8.76} & \textbf{9.03} & 6.52 & 7.03 & \textbf{7.30} & 7.21 & 7.03 & 6.61 & 7.06 & \textbf{7.30} & 7.12 & 7.03 & 7.70 \\
\ourmethod{} L3:1 & \textbf{8.91} & 9.24 & \textbf{9.33} & 8.67 & \textbf{9.03} & \textbf{6.97} & \textbf{7.24} & \textbf{7.30} & \textbf{7.24} & \textbf{7.18} & \textbf{6.88} & \textbf{7.21} & 7.21 & \textbf{7.18} & \textbf{7.12} & \textbf{7.79} \\
\ourmethod{} GT Text & 7.91 & 8.39 & 8.82 & 8.55 & 8.42 & 6.73 & 7.03 & 7.15 & 6.97 & 6.97 & 6.36 & 6.94 & 6.97 & 6.97 & 6.82 & 7.39 \\
% TextOnly & 3.30 & 3.30 & 3.28 & 2.82 & 3.18 & - & - & - & - & - & - & - & - & - & - & - \\
\bottomrule
\end{tabular}
}
% \vspace{-1.0\baselineskip} % Remove space after the table
\end{table*}

\section{Experiments}
In this section, we assess the effectiveness of \mbox{\ourmethod{}}, focusing on the semantic meaningfulness and the turn-taking behaviors of the generated dialogues.

\subsection{Experimental Setups}
\label{sec:experimentalsetups}
\subsubsection{Training Data}
We utilize the Fisher dataset \cite{fisher}, a double-channel telephone conversation corpus containing 2,000 hours of audio data, to train \ourmethod{}. Following prior works \cite{dgslm,syncllm}, we pre-process each recording in the Fisher dataset into segments of 120-second clips for training and evaluation. We divide the Fisher dataset into training, validation, and test sets, with detailed information provided in Appendix \ref{sec:datasetsplit}.
% Additionally, we incorporate the Candor dataset \cite{candor}, another double-channel conversation corpus, for evaluation purposes.

\subsubsection{Baselines and Our Approach}
We consider the following models or methods as our baselines.
\textbf{1) dGSLM \cite{dgslm}:} a dual-Transformer architecture, each modeling an individual audio channel.
\textbf{2) Speech Token Interleaving (STI):} We train the GLM-4-Voice backbone on the Fisher dataset by interleaving the speech tokens from the two channels, one token at a time.
\textbf{3) Speech Chunk Interleaving (SCI):} Similar to Speech Token Interleaving, but it interleaves speech tokens chunk by chunk with a chunk size of 5.
\textbf{4) Text-speech Token Alignment (Moshi Training Strategy, Moshi TS \cite{moshi}):} The only existing method for text-guided speech generation in e2e FD-SLMs. It involves inserting each text token exactly at the time its corresponding sound is produced. However, Moshi is trained using a wide range of datasets, with Fisher being only one of them, making it not directly comparable to our Fisher-finetuned baselines. To address this, we train the GLM-4-Voice backbone on the Fisher dataset with Moshi's training methodology to compare it with other baselines. Our detailed implementation of Moshi TS is illustrated in Appendix \ref{sec:moshitsImplementation}.
\textbf{5) Moshi (7B):} We also include the original Moshi model for completeness.
\textbf{6) Our Approach (\ourmethod{}, 9B):} We train the GLM-4-Voice backbone on the Fisher dataset by utilizing the text-guided full duplex dialogue modeling framework introduced in Section \ref{sec:dialoguemodeling}. The text-speech dataset is pre-processed using the framework illustrated in Section \ref{sec:turnsegmentation}.
\begin{remark}
While there are other e2e FD-SLMs, most have not been (fully) open-sourced yet, as summarized in Table \ref{tab:opensource}. Therefore, we mainly evaluate our approach against dGSLM, Moshi, and our trained baselines.
\end{remark}
% \begin{remark}
% Most e2e FD-SLMs demonstrate effectiveness by showing superiority over dGSLM \cite{syncllm,ntpp,moshi}. Therefore, we use the same way to demonstrate that our trained baseline performs reasonably well.
% \end{remark}

\subsubsection{Model Training and Inference}
The models are trained on the Fisher dataset for 2 epochs with a global batch size of 256, a learning rate of 4e-6, a cosine learning rate scheduler, no weight decay, and 20 warmup steps. During inference, we sample 1,000 audio clips from the Fisher test sets. The first 30 seconds serve as prompts, while the model generates 90-second continuations using two methods: unconditional and conditional generation. In unconditional generation, we train the model with text guidance in both audio channels, and the model generates both channels for the 90-second continuation during inference. This setting has potential in two-speaker podcast generation. In conditional generation, we train the model with only assistant-channel text guidance, and the model generates the assistant-channel audio based on the user-channel audio as context, which can be used for spoken interaction purposes.
% \footnote{Hyperparameters are found via grid search.}

\subsection{Results}
\label{sec:results}
% NOTE: add GPT-4o evaluation after
We aim to answer four critical Research Questions (RQs) through comprehensive experiments.
% \begin{itemize}[left=0pt]
%     \item \textbf{RQ1:} Does our method significantly improve the semantic meaningfulness and coherence of the generated full-duplex dialogue?
%     \item \textbf{RQ2:} Can our method preserve the natural conversational flow after integrating additional texts into the sequence?
%     \item \textbf{RQ3:} How does the model perform if we insert text data with different timings and lengths?
% \end{itemize}
\textbf{RQ1:} Does our method significantly improve the semantic meaningfulness and coherence of the generated full-duplex dialogue?
\textbf{RQ2:} Can our method preserve the natural conversational flow after integrating additional texts into the sequence?
\textbf{RQ3:} How does the model perform if we insert text data with different timings and lengths?
\textbf{RQ4:} How well does \ourmethod{} preserve GLM-4-Voice's original QA abilities?

% Offline Latency Version
\begin{table*}[t]
    \centering
    \caption{Full-Duplex-Bench evaluation results. The best performance is bold, and the second-best performance is underlined. Latency is presented in seconds. The scores in brackets indicate temperature.}
    \setlength{\tabcolsep}{1.13mm}{
    % \small
    \scalebox{1.0}{
    \begin{tabular}{cccccccccc}
        \toprule
        \multicolumn{1}{c}{\textbf{Dimension}} & \multicolumn{1}{c}{\textbf{Pause Handling}} & \multicolumn{3}{c}{\textbf{Backchannel}} & \multicolumn{2}{c}{\textbf{Smooth Turn Taking}} & \multicolumn{3}{c}{\textbf{User Interruption}} \\
        \cmidrule(lr){2-2} \cmidrule(lr){3-5} \cmidrule(lr){6-7} \cmidrule(lr){8-10}
        \textbf{Data} & Candor & \multicolumn{3}{c}{ICC} & \multicolumn{2}{c}{Candor}  & \multicolumn{3}{c}{Synthetic} \\
        \textbf{Metric} & TOR ($\downarrow$) & TOR ($\downarrow$) & Freq ($\uparrow$) & JSD ($\downarrow$) & TOR ($\uparrow$) & Latency ($\downarrow$) & TOR ($\uparrow$) & GPT-4o($\uparrow$) & Latency ($\downarrow$) \\
        \midrule
        dGSLM & 0.935  & 0.691 & 0.015  & 0.934  &  \textbf{0.975} & 0.352  & \underline{0.917} & 0.201 & 2.531 \\
        Moshi & 0.980 & 1.000 & 0.001 & 0.957 & 0.941 & \underline{0.265} & \textbf{1.000} & 0.765 & \textbf{0.257} \\
        \ourmethod{} L2:1 (0.8) & \textbf{0.390} & 0.667 & 0.013 & 0.927 & 0.773 & 0.567 & 0.558 & 1.349 & 2.263 \\ 
        \ourmethod{} L2:1 (0.9) & \underline{0.441} & \underline{0.519} & 0.019 & \underline{0.860} & 0.773 & 0.357 & 0.689 & \underline{1.441} & 1.870 \\ 
        \ourmethod{} L2:1 (1.0) & 0.587 & 0.630 & \underline{0.021} & 0.887 & 0.807 & 0.282 & 0.866 & \textbf{1.471} & 1.566 \\ 
        \ourmethod{} L2:1 (1.3) & 0.789 & \textbf{0.500} & \textbf{0.043} & \textbf{0.814} & \underline{0.950} & \textbf{0.102} & \textbf{1.000} & 1.150 & \underline{0.900} \\ 
        % Freeze-Omni & \textbf{0.642} & \textbf{0.481} & \textbf{0.636} & 0.001 & 0.997 & 0.336 & 0.953 & 0.867 &\textbf{3.615} & 1.409 \\ 
        % \textcolor{gray}{Gemini Live} & \textcolor{gray}{0.255} & \textcolor{gray}{0.310} & \textcolor{gray}{0.091} & \textcolor{gray}{0.012} & \textcolor{gray}{0.896} & \textcolor{gray}{0.655} & \textcolor{gray}{1.301} & \textcolor{gray}{0.891} & \textcolor{gray}{3.376} & \textcolor{gray}{1.183} \\
        \bottomrule
    \end{tabular}}}
    % \vspace{-5mm}
    \label{tab:fullduplexbench}
    % \vspace{-1.0\baselineskip} % Remove space after the table
\end{table*}

\begin{table*}[t]
\centering
\caption{\label{tab:turntakingall}
Pearson correlation results ($\uparrow$) of corpus-level turn-taking events with unconditional and conditional generation of various FD-SLMs evaluated under different temperature configurations.
}
\scalebox{0.9}{
\begin{tabular}{lccccc|ccccc|ccccc|c} 
\toprule
\multirow{2}{*}{\textbf{SLM}} & \multicolumn{5}{c|}{\textbf{Occurrence}} & \multicolumn{5}{c|}{\textbf{Cumulative Duration}} & \multicolumn{5}{c|}{\textbf{Averaged Duration}} & \multirow{2}{*}{\textbf{Overall}} \\
\cmidrule(lr){2-6} \cmidrule(lr){7-11} \cmidrule(lr){12-16}
& \textbf{0.8} & \textbf{0.9} & \textbf{1.0} & \textbf{1.3} & \textbf{AVG} & \textbf{0.8} & \textbf{0.9} & \textbf{1.0} & \textbf{1.3} & \textbf{AVG} & \textbf{0.8} & \textbf{0.9} & \textbf{1.0} & \textbf{1.3} & \textbf{AVG} & \\
\midrule
\rowcolor{beaublue} \multicolumn{17}{c}{\textbf{Unconditional}} \\
dGSLM & 0.33 & 0.41 & 0.45 & 0.36 & 0.39 & 0.45 & 0.53 & 0.55 & 0.40 & 0.48 & 0.30 & 0.37 & 0.42 & 0.26 & 0.34 & 0.40 \\
STI & 0.40 & 0.47 & 0.43 & 0.46 & 0.44 & 0.39 & 0.37 & 0.38 & 0.48 & 0.41 & 0.25 & 0.22 & 0.23 & 0.27 & 0.24 & 0.36 \\
SCI & 0.57 & 0.57 & 0.61 & 0.62 & 0.59 & 0.48 & 0.48 & 0.52 & 0.58 & 0.51 & 0.26 & 0.25 & 0.33 & 0.36 & 0.30 & 0.47 \\
Moshi TS & 0.52 & 0.53 & 0.56 & 0.61 & 0.56 & 0.38 & 0.43 & 0.46 & 0.54 & 0.45 & 0.20 & 0.24 & 0.28 & 0.34 & 0.26 & 0.42 \\
\ourmethod{} & 0.57 & 0.57 & 0.63 & 0.62 & 0.60 & 0.46 & 0.52 & 0.53 & 0.56 & 0.52 & 0.27 & 0.25 & 0.34 & 0.36 & 0.30 & 0.47 \\
\ourmethod{} L2:1 & 0.57 & 0.60 & 0.62 & 0.62 & 0.60 & 0.50 & 0.51 & 0.55 & 0.56 & 0.53 & 0.29 & 0.26 & 0.36 & 0.37 & 0.32 & 0.48 \\
\ourmethod{} L3:1 & 0.58 & 0.59 & 0.63 & 0.58 & 0.60 & 0.47 & 0.50 & 0.54 & 0.55 & 0.51 & 0.31 & 0.25 & 0.33 & 0.37 & 0.32 & 0.47 \\
% \ourmethod{} w/o Punc & 0.57 & 0.58 & 0.61 & 0.62 & 0.59 & 0.49 & 0.50 & 0.53 & 0.55 & 0.51 & 0.27 & 0.32 & 0.31 & 0.35 & 0.31 & 0.47 \\
\midrule
\rowcolor{beaublue} \multicolumn{17}{c}{\textbf{Conditional}} \\
STI & 0.50 & 0.52 & 0.58 & 0.62 & 0.56 & 0.40 & 0.42 & 0.47 & 0.53 & 0.46 & 0.20 & 0.21 & 0.27 & 0.32 & 0.25 & 0.42 \\
SCI & 0.47 & 0.49 & 0.49 & 0.50 & 0.49 & 0.28 & 0.40 & 0.46 & 0.53 & 0.42 & 0.17 & 0.25 & 0.29 & 0.34 & 0.26 & 0.39 \\
Moshi & 0.65 & 0.62 & 0.55 & 0.51 & 0.58 & 0.43 & 0.43 & 0.46 & 0.35 & 0.42 & 0.32 & 0.29 & 0.35 & 0.33 & 0.32 & 0.44 \\
Moshi TS & 0.48 & 0.48 & 0.48 & 0.45 & 0.47 & 0.39 & 0.39 & 0.43 & 0.50 & 0.43 & 0.24 & 0.25 & 0.29 & 0.29 & 0.27 & 0.39 \\
\ourmethod{} & 0.46 & 0.50 & 0.51 & 0.46 & 0.48 & 0.38 & 0.46 & 0.48 & 0.50 & 0.45 & 0.24 & 0.26 & 0.34 & 0.32 & 0.29 & 0.41 \\
\ourmethod{} L2:1 & 0.50 & 0.48 & 0.51 & 0.43 & 0.48 & 0.39 & 0.49 & 0.54 & 0.49 & 0.48 & 0.27 & 0.33 & 0.33 & 0.33 & 0.31 & 0.42 \\
\ourmethod{} L3:1 & 0.46 & 0.48 & 0.49 & 0.47 & 0.48 & 0.39 & 0.46 & 0.51 & 0.50 & 0.46 & 0.20 & 0.30 & 0.28 & 0.32 & 0.27 & 0.41 \\
% \ourmethod{} w/o Punc & 0.46 & 0.49 & 0.49 & 0.49 & 0.48 & 0.39 & 0.44 & 0.50 & 0.50 & 0.46 & 0.25 & 0.29 & 0.34 & 0.27 & 0.29 & 0.41 \\
\bottomrule
\end{tabular}
}
% \vspace{-1.0\baselineskip} % Remove space after the table
\end{table*}

\begin{figure}[t]
    \centering
    \includegraphics[width=0.49\textwidth]{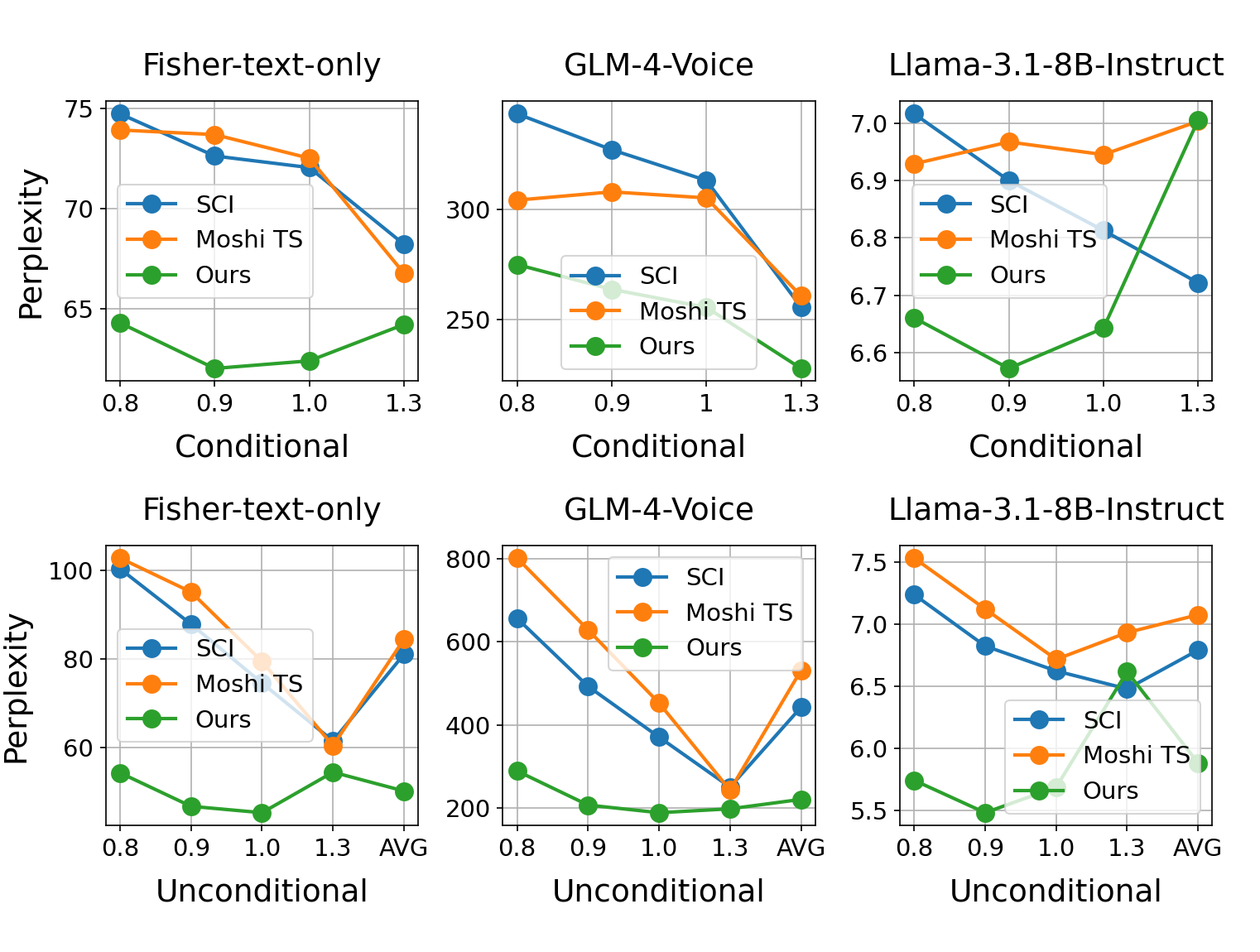}
    % \vspace{-9pt} % Remove space after the figure
    \caption{Perplexity evaluation of FD-SLMs across temperatures using Fisher-text-only, GLM-4-Voice, and Llama-3.1-8B-Instruct as evaluation models.}
    \label{fig:perplexityplot}
    % \vspace{-1.5\baselineskip} % Remove space after the figure
\end{figure}

\subsubsection{RQ1}
To evaluate the semantics of the generated dialogue, the most intuitive method is to conduct human evaluations. However, considering the time, cost, and potential inconsistencies associated with human evaluations, we propose utilizing the GPT-score. Specifically, given a generated double-channel dialogue audio, we first transcribe the audio from each channel into text with sentence-level timestamps. For this process, we employ the stable-ts package \cite{stable-ts} with the Whisper-large-v3 model. Subsequently, the transcribed sentences from both channels are grouped based on their start times, effectively representing the dialogue between the two speakers in text form along the timeline. Finally, the consolidated text is analyzed using GPT-4o \cite{gpt4osystemcard}, which assesses the meaningfulness and coherence of the dialogue on a scale of 0 to 5 with detailed rubrics. The complete GPT-4o prompts for unconditional and conditional generation settings are shown in Appendix \ref{sec:semanticEvalPrompts}. To demonstrate the reliability of GPT-based evaluation, we conduct human evaluations, showing strong consistency between human- and GPT-evaluated scores (see Section \ref{apx:humaneval}).

% \textbf{Baseline Comparison.} We compare our trained baselines, STI and SCI, with dGSLM in the unconditional generation setting on the Fisher test set (the same setting used in the dGSLM paper). As shown in Table \ref{tab:allbaseline}, STI and SCI perform significantly better than dGSLM, showing that our baselines are trained reasonably well.

We demonstrate the effectiveness of \ourmethod{} through evaluations in both unconditional and conditional settings. In the ``Conditional - All'' scenario, GPT-4o uses ordered transcripts from both channels to assess responses based on user speech. However, we observe that GPT-4o overly prioritizes user-assistant interactions, neglecting the coherence of assistant-generated speech within its channel. To address this, the ``Conditional - Assistant'' setting provides only the assistant-channel transcript, enabling a focused analysis of semantic quality within the assistant channel. We also observe that GPT-4o tends to assign conservative scores for this task, which is evidenced by the ground truth text from the Fisher dataset achieving an average GPT-score of 3.3. Therefore, we rescale all GPT-score results from 0-3.3 to \textbf{0-10.0}.

We compare \ourmethod{} with various baselines under different temperatures (as shown in Table \ref{tab:semanticeval}) and make the following observations. \textbf{1) Significant performance improvements over baselines.} \ourmethod{} performs significantly better than SCI and Moshi TS (over 30\% performance gain), showing that \ourmethod{} successfully leverages the text information to boost the meaningfulness and coherence of the generated dialogue. 
Notably, although Moshi is pre-trained with 10 times more real-world speech data compared to GLM-4-Voice\footnote{\textasciitilde7M hours versus \textasciitilde700k hours according to their papers}, \mbox{\ourmethod{}} achieves over 20\% performance gains, highlighting the effectiveness of our training strategy.
\textbf{2) Speech-only methods and text-speech methods excel at different temperatures.} SCI consistently performs the best at the largest temperature (1.3), whereas \ourmethod{} tend to perform the best at temperature 1.0. Additionally, \ourmethod{} shows more performance gain under lower temperatures, such as 0.8 and 0.9. \textbf{3) Focusing more on text generation makes the model perform better.} In ``\ourmethod{} L2:1" and ``\ourmethod{} L2:1" settings, we set the training loss weight of text and speech tokens to be 2:1 and 3:1, and the results show that they consistently outperform the standard 1:1 setting. \textbf{4) \ourmethod{} achieves performance comparable to its upper bounds.} In Table \ref{tab:semanticeval}, ``\ourmethod{} GT Text" refers to the GLM-4-Voice model trained with our proposed framework but switches to the ground truth transcriptions from the Fisher dataset. \ourmethod{} performs similarly with \ourmethod{} GT Text in the 1:1 loss setting (7.27 vs. 7.39), and the \ourmethod{} 2:1 and 3:1 outperform the upper bound in 1:1 loss.

\textbf{Further validation with perplexity.} To facilitate the semantic evaluation using GPT-score, perplexity is employed to further demonstrate the superior semantic abilities of \ourmethod{}. Specifically, we calculate perplexity using the same ordered transcript used in the GPT-score evaluation. Perplexity is measured with three distinct language models: Fisher-text-only, GLM-4-Voice, and Llama-3.1-8b-Instruct \cite{llama3}. The Fisher-text-only model, trained using Fisher-provided text transcript data on the GLM-4-Voice base model, serves as the ground truth for text-based dialogue distribution. GLM-4-Voice acts as the SLM backbone, while Llama-3.1-8b-Instruct serves as the text-only model backbone. We analyze the perplexity of generated dialogues across different temperature settings in both conditional and unconditional scenarios. As illustrated in Figure \ref{fig:perplexityplot}, \ourmethod{} consistently achieves lower perplexity compared to baseline approaches, demonstrating superior semantic modeling abilities.

\subsubsection{RQ2}
\label{sec:result_FDB}
Evaluating the naturalness of FD-SLM interaction involves assessing various turn-taking events in the full-duplex dialogues they generate. Traditional FD-SLM studies \cite{dgslm,syncllm,moshi} often analyze turn-taking behaviors at the corpus level, comparing the model-generated statistics to the corresponding ground-truth statistics. However, Lin et al. \cite{fullduplexbench} highlight that corpus-level statistics are often difficult to interpret, making them less suitable as an evaluation method. To overcome this limitation, they introduce the Full-Duplex-Bench benchmark, designed to evaluate the fine-grained turn-taking behaviors of FD-SLMs. This benchmark specifically assesses core turn-taking abilities, such as Smooth Turn-taking, User Interruption, Pause Handling, and Backchanneling \cite{fullduplexbench}.
% The detailed explanation of the four turn-taking events is in Appendix \ref{sec:FDBturntakingExplanations}.

We compare \ourmethod{} trained with 2:1 loss with the baseline approaches on Full-Duplex-Bench (as shown in Table \ref{tab:fullduplexbench}) and highlight the following. \textbf{1) \ourmethod{} demonstrates superior interaction naturalness compared to baseline approaches.} \ourmethod{} surpasses baselines on nearly all the metrics of the four turn-taking events, except for the TOR of smooth turn-taking and the latency of user interruption. \textbf{2) \ourmethod{} achieves optimal turn-taking performance when the temperature is set to 1.3.} At this temperature setting, \ourmethod{} outperforms not only the baselines but also \ourmethod{} at other temperatures. We note that \ourmethod{} is evaluated in offline mode, so the latencies shown in Table \ref{tab:fullduplexbench} do not represent the end-to-end streaming latency. This evaluation approach is consistent with the baseline results reported in Lin et al. \cite{fullduplexbench}, so the results are fully comparable. However, we acknowledge that the streaming latency of the chunk interleaving approaches (\ourmethod{}) is theoretically higher than token interleaving approaches (Moshi), and we provide the latency analysis and comparison in Section \ref{sec:latencyanalysis}.
% while \ourmethod{} is evaluated in offline mode, the italicized latency scores in blue are estimated streaming latencies calculated to ensure a fair comparison. A justification for offline inference and the conversion methodology is provided in Section \ref{sec:latencyanalysis}.

\textbf{Corpus-Level Turn-Taking Evaluations.} We also incorporate the results of corpus-level turn-taking evaluations. Following prior works, the turn-taking events considered here include Inter-Pausal Units (IPUs), pause, gap, and overlap \cite{dgslm}. Figure \ref{fig:turn_taking_events} in the supplementary material provides examples of these turn-taking events. To give a brief explanation of how these turn-taking events are extracted from the generated dialogue: First, the speech segments within each speaker channel are identified using a \textit{pyannote} VAD module. An IPU is defined as continuous speech within a single speaker channel. Multiple speech segments are grouped into a single IPU if the pause between them is shorter than $\tau_\text{IPU} $. As previously mentioned, $ \tau _ \text { IPU } $ is initially set to 0.5 seconds in the multi-modal turn segmentation and alignment framework (Section \ref{sec:turnsegmentation}). However, for the evaluations, it is adjusted to 0.2 seconds in accordance with prior studies \cite{dgslm,syncllm,ntpp}. A Pause refers to the absence of speech within the same speaker channel, while a Gap indicates the discontinuation of speech across different channels. Overlap refers to the time period during which both speakers are speaking simultaneously.

We calculate the Pearson correlation of the occurrence, cumulative duration, and average duration of the turn-taking events between the 30-second prompts and the corresponding 90-second continuations, as described in \cite{syncllm, dgslm}. This measures whether the model can generate dialogues with similar conversational features to the prompt. The metric is averaged across the four turn-taking events.

The corpus-level turn-taking evaluation results on the Fisher testset are presented in Table \ref{tab:turntakingall}. We conduct evaluations under both unconditional and conditional settings, comparing \ourmethod{} with various baseline approaches. The results indicate that \textbf{\ourmethod{} demonstrates comparable performance to the baselines} in terms of Pearson correlation for occurrence, cumulative duration, and average duration. This finding suggests that \ourmethod{}, which integrates additional text streams into the modeling process, does not exhibit significant behavioral differences at the corpus level.

% \begin{remark}
% We include the corpus-level turn-taking results for the sake of completeness. However, as noted in Section \ref{sec:result_FDB}, corpus-level statistics are less interpretable and not as suitable for evaluating the turn-taking abilities of FD-SLMs. Therefore, we recommend readers focus on the more reliable results assessed using the Full-Duplex-Bench benchmark (Section \ref{sec:result_FDB}).
% \end{remark}

% \begin{table}[t]
% \centering
% \scalebox{0.9}{
% \begin{tabular}{lccccc} 
% \toprule
% \multirow{2}{*}{\textbf{SLM}} & \multicolumn{5}{c}{\textbf{Conditional - All}} \\
% \cmidrule(lr){2-6} 
% & \textbf{0.8} & \textbf{0.9} & \textbf{1.0} & \textbf{1.3} & \textbf{AVG} \\
% \midrule
% \ourmethod{} & \textbf{2.06} & \textbf{2.17} & \textbf{2.34} & \textbf{2.36} & \textbf{2.23} \\
% Long & 1.89 & 1.97 & 2.08 & 2.15 & 2.02 \\
% Short & 2.00 & 2.15 & 2.22 & 2.21 & 2.15 \\
% Early & 1.61 & 1.67 & 1.86 & 2.27 & 1.85 \\
% Late & 1.74 & 1.90 & 2.00 & 2.21 & 1.96 \\
% \bottomrule
% \end{tabular}
% }
% \caption{\label{tab:longshortearlylate}
% Semantic evaluation results of \ourmethod{} on different text insertion timings and lengths.
% }
% \vspace{-1.0\baselineskip} % Remove space after the table
% \end{table}

\begin{table}[t]
\centering
\caption{\label{tab:longshortearlylate}
Semantic evaluation results ($\uparrow$) of \ourmethod{} on different text insertion timings and lengths.
}
\scalebox{1.0}{
\begin{tabular}{lccccc} 
\toprule
\multirow{2}{*}{\textbf{SLM}} & \multicolumn{5}{c}{\textbf{Conditional - All}} \\
\cmidrule(lr){2-6} 
& \textbf{0.8} & \textbf{0.9} & \textbf{1.0} & \textbf{1.3} & \textbf{AVG} \\
\midrule
\ourmethod{} & \textbf{6.24} & \textbf{6.58} & \textbf{7.09} & \textbf{7.15} & \textbf{6.76} \\
Too Long & 5.73 & 5.97 & 6.30 & 6.52 & 6.12 \\
Too Short & 6.06 & 6.52 & 6.73 & 6.70 & 6.52 \\
Too Early & 4.88 & 5.06 & 5.64 & 6.88 & 5.61 \\
Too Late & 5.27 & 5.76 & 6.06 & 6.70 & 5.94 \\
\bottomrule
\end{tabular}
}
% \vspace{-1.0\baselineskip} % Remove space after the table
\end{table}

\begin{table}[t]
  \centering
  \caption{\label{tab:qabenchmarks}
    Comparison of model performance (S$\rightarrow$T, $\uparrow$) across QA benchmarks utilized in GLM-4-Voice.
  }
  % \rowcolors{1}{white}{lightblue}
  % \small % or \footnotesize or \scriptsize
\scalebox{0.82}{
  \begin{tabular}{lccc} % Adjust the width as needed
    \toprule
    \textbf{Benchmarks} & \textbf{Web Questions} & \textbf{Llama Questions} & \textbf{Trivia QA} \\
    \midrule
    \textbf{SpeechGPT}    & 6.5  & 21.6 & 14.8 \\
    \textbf{Spectron}     & 6.1  & 21.9 & --   \\
    \textbf{Moshi}        & 26.6 & 62.3 & 22.8 \\
    \textbf{GLM-4-Voice}  & 32.2 & 64.7 & 39.1 \\
    \textbf{\ourmethod{} L2:1} & 32.5 & 64.0 & 32.0 \\
    \bottomrule
  \end{tabular}
}
\end{table}

\subsubsection{RQ3}
We aim to support the claim that the text insertion in the full duplex dialogue modeling process exhibits the insertion timing and length challenges. Specifically, we simulate the following four text insertion scenarios and demonstrate the superiority of our approach. \textbf{1) Too Long:} Each text chunk contains 20 text tokens. \textbf{2) Too Short:} Each text chunk contains 2 text tokens. \textbf{3) Too Early:} The insertion time of every text chunk is shifted to 4s earlier. \textbf{4) Too Late:} The insertion time of every text chunk is shifted to 4s later.

The results presented in Table \ref{tab:longshortearlylate} highlight the following. \textbf{1) Text insertion timings and lengths indeed pose challenges to the full duplex dialogue modeling process.} We observe that all the simulated settings significantly degrade the model performance. \textbf{2) Model performance is particularly sensitive to insertion timing.} The \textit{Too Early} and \textit{Too Late} scenarios result in greater performance degradation compared to \textit{Too Long} and \textit{Too Short}. This indicates that accurate text-speech alignment is crucial in full duplex dialogue modeling.

% The results presented in Table \ref{tab:longshortearlylate} emphasize the following key findings. \textbf{1) The timing and length of text insertions pose significant challenges to the full-duplex dialogue modeling process.} Our analysis shows that all simulated conditions lead to notable degradation in model performance. \textbf{2) Model performance is particularly sensitive to insertion timing.} Scenarios involving "Too Early" and "Too Late" insertions result in greater performance degradation compared to the "Too Long" and "Too Short" scenarios.

\subsubsection{RQ4}
We utilize the three widely used Speech QA benchmarks to evaluate how much \ourmethod{} preserves the original GLM-4-Voice's QA abilities. All the baseline results are copied from GLM-4-Voice \cite{glm4voice}. As presented in Table \ref{tab:qabenchmarks}, \ourmethod{}'s performance on the three benchmarks is comparable to GLM-4-Voice. This demonstrates that TurnGuide successfully preserves most of the original model's QA abilities.

\begin{remark}
We use the half-duplex inference mode of \ourmethod{} (same as GLM-4-Voice) due to the lack of QA data in our training set. We observe that \ourmethod{} still generates Fisher-styled chitchat when prompted with Speech QA data in the full duplex inference mode.
\end{remark}

\begin{table}[t]
  \centering
  \caption{\label{tab:listeningtest}
    Alignment accuracy ($\uparrow$) between human judgments and GPT-4o semantic evaluation scores for generated dialogue pairs with varying score differences.
  }
  % \rowcolors{1}{white}{lightblue}
  % \small % or \footnotesize or \scriptsize
\scalebox{1.0}{
  \begin{tabular}{lcccc} % Adjust the width as needed
    \toprule
    \textbf{Difference}  & \textbf{1}  & \textbf{2} & \textbf{3} & \textbf{Overall} \\
    \midrule
    % \rowcolor{lightblue} \multicolumn{4}{c}{\textbf{Direct Answer (String-matching)}} \\ % Single word centered across all columns
    Accuracy       & 7/10 & 9/10 & 9/10 & 25/30 \\
    \bottomrule
  \end{tabular}
}
\end{table}

\begin{table}[t]
  \centering
  \caption{\label{tab:listeningtest_naturalness}
    Absolute rating results ($\uparrow$) of the human evaluations assessing the Meaningfulness (M-MOS) and Naturalness (N-MOS) of the dialogues.
  }
  % \rowcolors{1}{white}{lightblue}
  % \small % or \footnotesize or \scriptsize
\scalebox{1.0}{
  \begin{tabular}{lcc} % Adjust the width as needed
    \toprule
    \textbf{SLMs}  & \textbf{M-MOS}  & \textbf{N-MOS} \\
    \midrule
    % \rowcolor{lightblue} \multicolumn{4}{c}{\textbf{Direct Answer (String-matching)}} \\ % Single word centered across all columns
    SCI       & 2.24 & 2.47 \\
    Moshi       & 2.69 & 2.75 \\
    \ourmethod{}       & \textbf{3.26} & \textbf{3.23} \\
    \midrule
    Ground Truth       & 4.25 & 4.29 \\
    \bottomrule
  \end{tabular}
}
\end{table}

\section{Discussion}
\subsection{Judgment Alignment between Human and GPT-4o in Semantic Evaluation}
\label{apx:humaneval}
In this paper, we primarily utilize GPT-4o to evaluate the semantics of the generated dialogues. This raises a crucial question: \textbf{Can AI evaluation scores accurately capture human preferences?} To investigate this, we conduct two human listening tests, including \textit{pairwise comparison} and \textit{absolute rating}, to examine the alignment between GPT-evaluated scores and human judgments.

% Since the semantic evaluation results heavily rely on the judgment of GPT-4o, it is essential to demonstrate the reliability of the LLM-based evaluation. To assess this, we conduct two human listening tests, including a \textit{pairwise comparison} and an \textit{absolute rating}, to examine the alignment between GPT-evaluated scores and human judgments.

\begin{itemize}[left=0pt]
    \item \textbf{Pairwise Comparison.} We evaluate GPT-4o’s ability to differentiate dialogue quality by randomly selecting 30 dialogue pairs with varying GPT-score margins (10 pairs each for differences of 1, 2, and 3). Three judges, provided with both audio and Whisper-large-v3 transcripts, assessed which dialogue in each randomized pair exhibited superior semantic quality. The results in Table \ref{tab:listeningtest} demonstrate strong human-AI alignment, with matching judgments in 25 out of 30 pairs. While alignment was slightly lower for pairs with minor score differences (7/10 for a 1-point difference), the high overall correlation underscores the reliability of GPT-based scoring for distinguishing semantic nuances.
    \item \textbf{Absolute Rating.} To further verify model rankings, nine judges performed an absolute rating task using a 5-point Likert scale, following the methodology of Veluri et al. \cite{syncllm}. We sampled 40 continuations (10 per model) from SCI, Moshi, \ourmethod{}, and the Ground Truth at their optimal temperatures. Evaluators rated each sample for Meaningfulness (M-MOS) and Naturalness (N-MOS). As shown in Table \ref{tab:listeningtest_naturalness}, the human-assigned rankings perfectly align with the automated rankings in Table \ref{tab:semanticeval}. Notably, \ourmethod{} outperformed all baseline models across both metrics, confirming its effectiveness in semantic modeling.
\end{itemize}

\subsection{Latency Analysis.}
\label{sec:latencyanalysis}
We theoretically analyze the overall interaction latency of \ourmethod{}, which represents the time taken for the user to hear the first output audio sample of the response. Since our model is a fine-tuned version of GLM-4-Voice, much of the latency calculation can refer to the original GLM-4-Voice paper. Specifically, the e2e latency is divided into three components: Speech Tokenization, LLM processing (including pre-filling and token decoding), and Speech Decoding.

With \ourmethod{} processing speech in chunks, the latency corresponds to the total processing time (Speech Tokenization + LLM + Speech Decoding) for the first chunk. Using a speech chunk size of five tokens, the latency is calculated as follows: 1) The Speech Tokenization time for 5 tokens ($T_{st}$) is 0.01s, the LLM processing time ($T_{LLM}$), which includes pre-filling 5 user speech tokens along with special tokens and decoding a total of 10 assistant tokens (5 text tokens + 5 speech tokens), is 0.3s, and the Speech Decoding time ($T_{SD}$) for 5 speech tokens is 0.38s. Thus, the total latency is $T_{ALL} = T_{st} + T_{LLM} + T_{SD} = 0.69s$. It is worth noting that the optimal speech decoding process involves decoding only 5 speech tokens, but the original GLM-4-Voice's vocoder needs to decode at least 10 tokens together to ensure acoustic quality. Therefore, reaching the optimal latency in our analysis requires fine-tuning the vocoder, but we consider this to be out of scope of this paper, as our main contribution is the data preparation and techniques for LLM training. Additionally, it is important to note that chunk interleaving methods generally have higher theoretical latency than token interleaving methods. For instance, \ourmethod{} processes five tokens simultaneously, while Moshi processes only one, resulting in a theoretical latency of 0.16s for Moshi. Nevertheless, \ourmethod{} significantly enhances semantic intelligence while maintaining a latency of under 0.7s---a usually acceptable latency for real-time spoken interactions \cite{freezeomni,omniflatten}.

% \textcolor{red}{We illustrate how the latencies of \ourmethod{} are derived in Table \ref{tab:fullduplexbench}. First, we conduct offline inference due to the reduced acoustic quality of the original GLM-4-Voice vocoder otherwise. Second, the streaming latency is essentially the original measured latency plus $T_{ALL}$. Thus, for each generated dialogue in Full-Duplex-Bench, we manually add 0.69 seconds to the measured latency and report the average in Table \ref{tab:fullduplexbench}. We note that some latencies for TurnGuide are lower than 0.69 seconds, which is due to some dialogues exhibiting negative latency. This occurs because the training dataset we use (i.e., Fisher) includes natural human-human speech, which inherently contains negative latency cases. Therefore, this occurrence is a normal behavior of our model rather than a design flaw.}

% Since our text-speech interleaving approach introduces up to five additional text tokens alongside the normal speech generation process, the only extra latency compared to the original GLM-4-Voice is the decoding time

\section{Conclusion}
We present \ourmethod{}, a novel method to improve the dialogue generation abilities of e2e FD-SLMs. By segmenting assistant speech into turns and training the model with channel-wise interleaving and turn-level text-speech interleaving strategies, our approach effectively addresses key challenges of text insertion timing and length, ensuring coherent and natural dialogue generation. Experiments show significant improvements of our method in semantic quality and various natural turn-taking events, validating its effectiveness for spoken dialogue systems and paving the way for more meaningful and human-like speech interactions.

\section{Limitations}
% Despite the advancements of \ourmethod{}, several limitations remain. First, while the Fisher dataset is a widely used dataset, its telephony-based nature may not fully capture the stylistic diversity of real-world conversational scenarios. Second, while our GPT-4o-based semantic metrics strongly align with human judgment in our listening tests, they may not capture every nuance of dialogue quality. Finally, although our analysis identifies an optimal decoding length of 5 speech tokens for latency, the current GLM-4-Voice vocoder requires at least 10 tokens to maintain acoustic stability; we consider the vocoder optimization beyond the scope of this work.

While the paper introduces \ourmethod{}, a novel dynamic turn-level text-speech interleaved generation approach for enhancing e2e FD-SLMs, it still presents several limitations. First, the effectiveness of \ourmethod{} is primarily validated on the Fisher dataset, which, although widely used, may not fully capture the diversity and complexity of real-world conversational scenarios since Fisher is a telephone conversation dataset. However, addressing this limitation would necessitate the effort of open-sourcing real-world double-channel dialogue datasets that encompass a broader range of styles. Second, the evaluation relies heavily on GPT-4o-based automated metrics (GPT-score) for semantic assessment, which, despite being aligned with human judgments in most cases, may not perfectly reflect the nuanced human perceptions of dialogue quality. Finally, although our analysis identifies an optimal decoding length of 5 speech tokens for latency, the current GLM-4-Voice vocoder requires at least 10 tokens to maintain acoustic stability; we consider the vocoder optimization beyond the scope of this work.

\section{Acknowledgements}
The research presented in this paper was partially supported by the Research Grants Council of the Hong Kong Special Administrative Region, China (CUHK 2300246, RGC C1043-24G), (CUHK 14203425, RGC GRF 2151317), and CUHK 4055263.

% \section{Use of Generative AI Disclosure}
% We use GPT-4o \cite{gpt4osystemcard} to polish the writing of the paper.

\bibliographystyle{IEEEtran}
\bibliography{mybib}

\appendix

% This forces all pending figures to render and starts a new page
\clearpage

\begin{figure*}[t]
    \centering
    \includegraphics[width=0.995\textwidth]{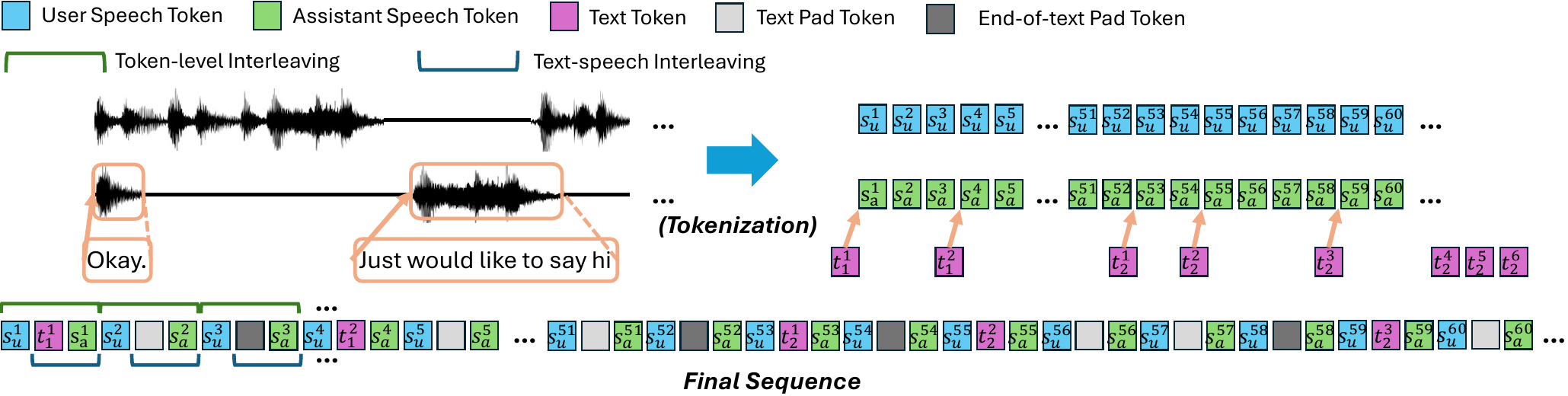}
    \caption{Illustration of the Moshi training strategy.}
    \label{fig:moshi_ts}
\end{figure*}

\begin{figure}[t]
    \centering
    \includegraphics[width=0.48\textwidth]{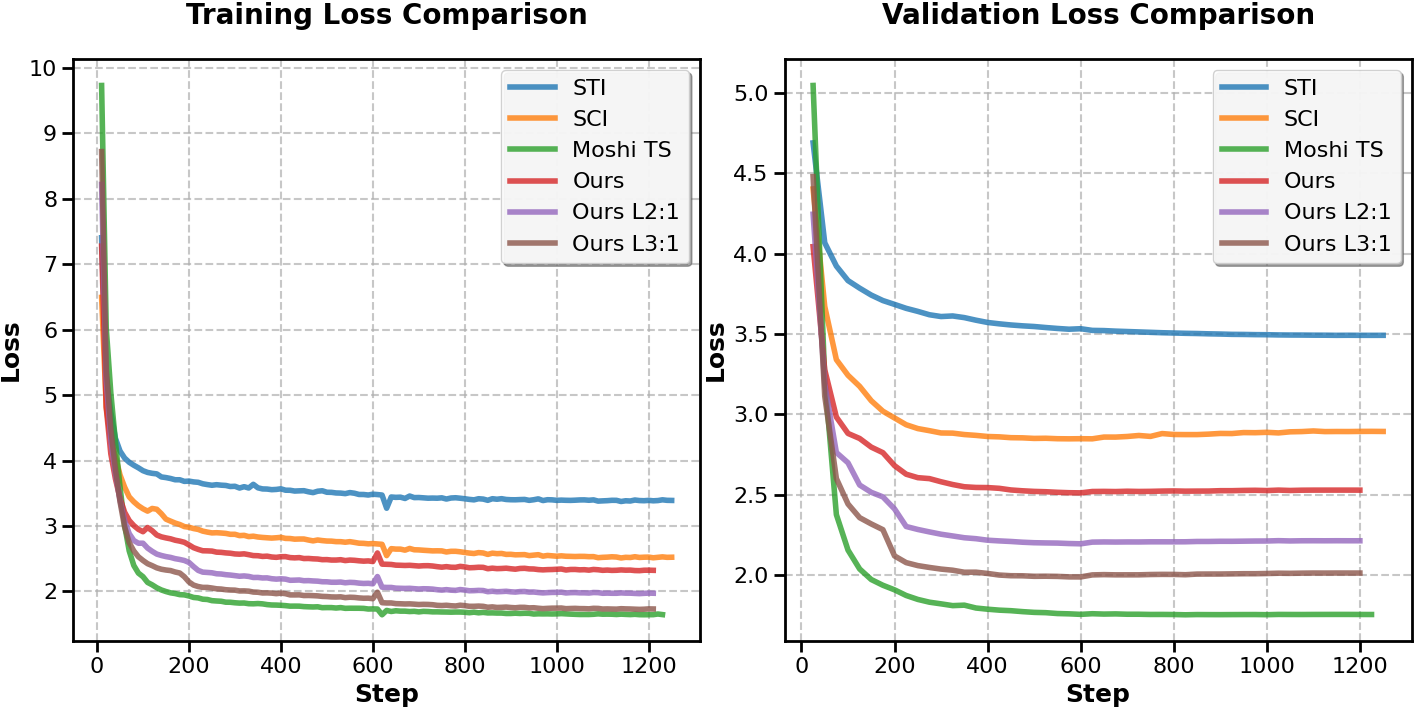}
    \caption{Training and validation loss comparison of different training strategies.}
    \label{fig:loss_compare}
\end{figure}

\begin{figure}[t]
    \centering
    \includegraphics[width=0.475\textwidth]{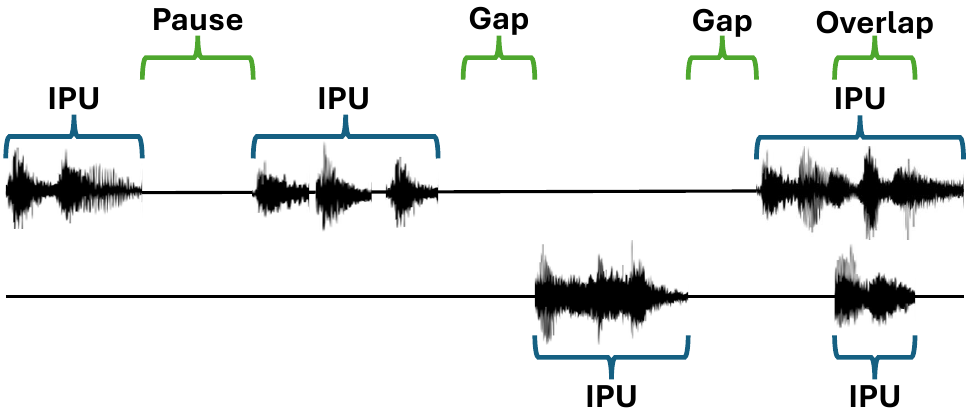}
    \caption{Illustration of the four turn-taking events used in the evaluation.}
    \label{fig:turn_taking_events}
\end{figure}

\section{Moshi TS Implementation}
\label{sec:moshitsImplementation}
As detailed in Section \ref{sec:experimentalsetups}, we do not directly compare our model with Moshi, as Moshi is a general-purpose SLM that utilizes different pre-training speech data and is not fine-tuned on the Fisher dataset during its final training stage. Instead, we opt to train a model using Moshi's training strategy, employing GLM-4-Voice on the Fisher dataset. This approach ensures a fair comparison, as the base model and training dataset remain identical, with the only difference between our approach and Moshi TS being the training strategy used. Therefore, in this section, we illustrate the details of the training process of Moshi TS.

First, Moshi utilizes eight codebooks, where every time step includes eight speech tokens for each channel. We adapt this technique for the GLM-4-Voice backbone, which relies on a single codebook method, allotting one speech token per channel per time step. Then, the tokens from each channel are interleaved at each time step, making Moshi rely on a token-level interleaving approach. In the original implementation of Moshi, a text-channel token precedes each assistant-channel speech token, which could either be a genuine text token, a text pad token, or an end-of-text pad token. Moshi also employs text-speech interleaving at token-level, where each text token is incorporated at the precise moment it is spoken. Consequently, if a time step features an audible text token, the corresponding text token is inserted. If not, either a text pad token or an end-of-text pad token is inserted, determined by whether the subsequent text-channel token is an actual text token. We follow the original Moshi implementation by extracting the timestamp of each word with \textit{Whisper-timestamped} package \cite{whisper-timestamped} and the medium Whisper model \cite{whisper}. The illustration of our implemented Moshi TS is shown in Figure~\ref{fig:moshi_ts}.

\section{Comparison of Different Training Strategies}
In this section, we provide a comparison of different training strategies and share corresponding insights. We start by examining the impact of these strategies on the training process by analyzing the training loss curves displayed in Figure \ref{fig:loss_compare}. First, \textbf{SCI achieves a significant reduction in training loss compared to STI.} This improvement can likely be attributed to the addition of speaker ID tokens for each channel, which provides extra information about the correspondence between speech tokens and their respective speakers, thereby simplifying the model's learning process. Furthermore, the reduction in training loss may also result from the organization of speech tokens into chunks, which enhances contextual coherence compared to STI. Second, \textbf{\ourmethod{} demonstrates even lower training loss than STI.} More specifically, \ourmethod{} outperforms both SCI and STI, with configurations such as \ourmethod{} L2:1 and \ourmethod{} L3:1 achieving even greater reductions in training loss. These results highlight the benefits of the text-speech alignment process, which facilitates SLM training and enables superior semantic performance. Finally, \textbf{the loss reduction observed in Moshi TS does not translate into performance improvements.} While Moshi TS achieves one of the lowest training losses, its overall performance trails behind our approach. This discrepancy may stem from the excessive presence of text pad tokens relative to actual text tokens. The repetitive nature of pad tokens simplifies prediction, leading to artificially low training losses without corresponding gains in real-world performance.

We then intuitively explore the reasons behind the lower performance of Moshi TS compared to \ourmethod{} and identify three key factors contributing to this difference. First, \textbf{Moshi TS tends to fragment text semantics.} Both Moshi TS and our method aim to enhance the semantics of generated speech using textual information. However, Moshi TS splits text tokens so significantly that it hampers the model's ability to learn coherent text representations. In contrast, \ourmethod{} aggregates these text tokens, creating a more cohesive grouping that simplifies the learning process for text generation, which leads to notable performance improvements. Second, \textbf{Moshi TS fails to mitigate the inherently fragmented nature of speech semantics.} Here, we propose a new way of understanding the semantic information within a speech audio. Speech audio combines multiple types of information together, including semantic, paralinguistic, and speaker information \cite{SLMsurvey}, and the semantics within speech can be defined by its transcription, with speech containing semantic information only at the temporal points when each word is sounded. This fragmented nature increases the difficulty for coherent speech generation. Although Moshi TS incorporates text into the speech generation process, it inserts text tokens strictly at the time they are sounded in the speech. This approach only supplements speech with semantic information at points where semantic information already exists, offering limited potential for enhancing semantic representation. By comparison, \ourmethod{} groups texts within the same turn and processes them collectively. This method introduces denser semantic information, effectively guiding the speech generation process to improve coherence. Third, \textbf{Moshi TS heavily depends on precise word-level timestamps from the ASR systems.} In practice, these timestamps are not always exact, often being slightly ahead or behind the actual sounding moment, which can disrupt text-speech alignment and degrade performance. \ourmethod{}, however, leverages only the start time of each text chunk, significantly reducing reliance on word-level timestamp accuracy. This simpler alignment strategy not only lowers the risk of errors but also contributes to stronger overall performance.

\section{The Fisher Dataset Split}
% \section{Fisher and Candor Dataset Split}
\label{sec:datasetsplit}
The Fisher dataset is frequently referenced and employed in e2e FD-SLM studies. However, to date, there is no official training, validation, and test set split for the dataset, nor do existing papers provide details of their dataset splits. This poses a significant challenge for future research, as it becomes difficult to replicate the reported model performance in original studies. Specifically, when researchers use their own dataset splits to train models and subsequently compare their model performance with publicly available checkpoints from prior work, there is a risk of test set leakage. This occurs when an existing model has been trained on test samples included within a new split defined by researchers. To tackle this issue, we propose releasing our train/validation/test splits for the Fisher dataset and strongly encourage future studies to adopt these splits for a more equitable and reliable basis of comparison. Specifically, Fisher separates all the audio into 117 parts (000-116). We split the train/validation/test set using part IDs 000-110/111-113/114-116, and we randomly select roughly 1,000 samples in our Fisher test set to perform all the evaluations reported in this paper. We will release the dataset split after the acceptance of the paper.
% Both the Fisher and Candor datasets are frequently referenced and employed in e2e FD-SLM studies. However, to date, there is no official training, validation, and test set split for these datasets, nor do existing papers provide details of their dataset splits. This poses a significant challenge for future research, as it becomes difficult to replicate the reported model performance in original studies. Specifically, when researchers use their own dataset splits to train models and subsequently compare their model performance with publicly available checkpoints from prior work, there is a risk of test set leakage. This occurs when an existing model has been trained on test samples included within a new split defined by researchers. To tackle this issue, we propose releasing our train/validation/test splits for the Fisher and Candor datasets and strongly encourage future studies to adopt these splits for a more equitable and reliable basis of comparison. Specifically, Fisher separates all the audio into 117 parts (000-116). We split the train/validation/test set using part IDs 000-110/111-113/114-116, and we randomly select roughly 1,000 samples in our Fisher test set to perform all the evaluations reported in this paper. For the Candor dataset, we randomly select roughly 1,000 samples from the part 001 audio data to perform the evaluations. We will report the IDs of the 1,000 samples for both datasets after the acceptance of the paper.

\section{Detailed Computing Infrastructure and Hyperparameter Specification}
\label{sec:detailedspecification}
Our training environment uses Pytorch 2.0.0 with the transformers library of version 4.49.0. Our inference environment uses Pytorch 1.5.0 with the transformers library of version 4.44.2. We use grid search \cite{gridsearch} for hyperparameter tuning, and we run each experiment once.

% We perform all the experiments on A800 GPUs, which have 80GB of GPU memory. The CPU model of the machine is Intel(R) Xeon(R) Platinum 8378A CPU @ 3.00GHz. The operating system is Ubuntu 20.04.4 LTS. Our training environment uses Pytorch 2.0.0 with the transformers library of version 4.49.0. Our inference environment uses Pytorch 1.5.0 with the transformers library of version 4.44.2. The training process requires 48 hours on 8 A800 GPUs. We use grid search \citep{gridsearch} for hyperparameter tuning, and we run each experiment once.

% \section{Case Study}
% \textcolor{red}{TBW}

\section{Semantic Evaluation Prompts}
\label{sec:semanticEvalPrompts}
We give prompts for the three settings of the GPT-score semantic evaluation used in Table \ref{tab:semanticeval}: We use the prompt in Figure \ref{fig:gptunconditionalprompt} for the unconditional setting, and use the prompt in Figure \ref{fig:gptconditionalprompt} for the two conditional settings.

\section{Explanations of Turn-Taking Events in Full-Duplex-Bench}
\label{sec:FDBturntakingExplanations}
This section details the four turn-taking events evaluated in the Full-Duplex-Bench. Specifically, \textbf{1) Smooth Turn-taking} evaluates the model’s ability to seamlessly take over the conversation after the user’s speaking turn. \textbf{2) User Interruption} assesses the model’s capability to recognize and adapt to new content introduced by the user during the interaction. \textbf{3) Pause Handling} determines whether the model can remain silent during brief pauses by the user before they complete their speaking turn. \textbf{4) Backchanneling} measures the model’s ability to provide brief acknowledgments at appropriate moments while the user is speaking. For a detailed explanation of the evaluation metrics, please refer to their original paper \cite{fullduplexbench}.

\section{Potential Risks}
The primary ethical concern regarding \ourmethod{} involves \textbf{safety and alignment risks} arising from the model's training data and the absence of post-hoc behavioral tuning. Specifically, since the model is trained on the Fisher dataset---a collection of real-world telephone conversations---it may inherit and replicate biased, sensitive, or harmful topics present in authentic human dialogue. These risks are compounded by the fact that the current iteration has not undergone Reinforcement Learning from Human Feedback (RLHF) \cite{rlhf_book,instructgpt,dpo} or similar safety alignment procedures to ensure its outputs remain consistent with human preferences. Without a robust refusal mechanism or a safety-oriented backbone, the model's enhanced capabilities for real-time interruption and overlapping speech could potentially be exploited to generate deceptive or inappropriate content. Consequently, \ourmethod{} is intended strictly for research purposes and is not suitable for production deployment without the integration of rigorous safety-tuning frameworks and content moderation safeguards.

\begin{figure*}[ht]
    \centering
    \begin{lstlisting}[style=mystyle]
Please evaluate the following two-speaker dialogue transcript for how meaningful the speech is (based on its content), focusing primarily on the part from 30 seconds to 120 seconds of the transcript. Use the following scale:

0: Completely meaningless; no coherent sentences, random words, or unintelligible.
0.5: Almost no meaning; isolated words or phrases, but no understandable ideas.
1: Extremely low meaning; rare, vague fragments of ideas, but mostly incoherent or off-topic.
1.5: Very little meaning; a few short, unclear ideas, but mostly disjointed or confusing.
2: Low meaning; some recognizable ideas or topics, but mostly unclear, incomplete, or off-topic.
2.5: Somewhat low meaning; a few coherent points, but overall lacks clarity or logical flow.
3: Moderate meaning; general topic is understandable, but there are gaps, unclear parts, or weak connections.
3.5: Fairly meaningful; mostly coherent and relevant, but with some confusion, repetition, or lack of detail.
4: Meaningful; clear and logical, with relevant and connected ideas, though may lack depth or detail.
4.5: Very meaningful; almost fully coherent, with well-developed, relevant, and connected ideas.
5: Extremely meaningful; highly coherent, clear, and detailed, with all ideas well connected and relevant.

Only output the final score (0, 0.5, 1, 1.5, ..., 5) **ONLY** according to the above rubric. Do not output anything else.
    \end{lstlisting}
    \caption{The prompt for GPT-score semantic evaluation under the unconditional setting.}
    \label{fig:gptunconditionalprompt}
\end{figure*}

\begin{figure*}[ht]
    \centering
    \begin{lstlisting}[style=mystyle]
Please evaluate the following two-speaker dialogue transcript for how meaningful the speech is (based on its content), only focusing on the model channel's output and from 30 seconds to 120 seconds of the transcript. Use the following scale:

0: Completely meaningless; no coherent sentences, random words, or unintelligible.
0.5: Almost no meaning; isolated words or phrases, but no understandable ideas.
1: Extremely low meaning; rare, vague fragments of ideas, but mostly incoherent or off-topic.
1.5: Very little meaning; a few short, unclear ideas, but mostly disjointed or confusing.
2: Low meaning; some recognizable ideas or topics, but mostly unclear, incomplete, or off-topic.
2.5: Somewhat low meaning; a few coherent points, but overall lacks clarity or logical flow.
3: Moderate meaning; general topic is understandable, but there are gaps, unclear parts, or weak connections.
3.5: Fairly meaningful; mostly coherent and relevant, but with some confusion, repetition, or lack of detail.
4: Meaningful; clear and logical, with relevant and connected ideas, though may lack depth or detail.
4.5: Very meaningful; almost fully coherent, with well-developed, relevant, and connected ideas.
5: Extremely meaningful; highly coherent, clear, and detailed, with all ideas well connected and relevant.

Only output the final score (0, 0.5, 1, 1.5, ..., 5) **ONLY** according to the above rubric. Do not output anything else.
    \end{lstlisting}
    \caption{The prompt for GPT-score semantic evaluation under the two conditional settings.}
    \label{fig:gptconditionalprompt}
\end{figure*}

\begin{figure*}[ht]
    \centering
    \begin{lstlisting}[style=mystyle]
## Pairwise comparison instructions

### Task overview
You will be shown **30 pairs** of two-speaker dialogue samples generated by different systems.
For each pair, your job is to choose **which dialogue has better semantic quality** (i.e., which one is more meaningful and coherent in what the speakers say).

### What you will receive for each pair
For each pair, you will be given:
- **Audio** of Dialogue A and Dialogue B.
- **Transcripts** for both dialogues (automatically generated with Whisper-large-v3 using the stable-ts package).
The order (A/B) is **randomized** for every pair; do not assume A or B is better.

### What "better semantic quality" means
When deciding which dialogue is better, focus on **content meaning and coherence**, not just fluency.
Prefer the dialogue that is:
- More **understandable** and stays on a recognizable topic.
- More **coherent**, with logical connections between utterances.
- Less **contradictory**, less nonsensical, and less off-topic.
- More **informative/complete** rather than fragmented or empty.

### What to ignore (or de-emphasize)
Do **not** primarily judge based on:
- Audio naturalness (voice quality, prosody, accent, loudness) unless it prevents understanding.
- Turn-taking style (interruptions/overlap) unless it makes the content hard to follow.
- Minor transcription errors, as long as the intended meaning is still clear from audio + context.

### How to make your choice
For each pair, listen to both dialogues (you may replay) and read transcripts as needed.
Then select exactly one option:
- **A is better** (Dialogue A has higher semantic quality).
- **B is better** (Dialogue B has higher semantic quality).

### Consistency and independence
Judge each pair **independently**; do not try to keep scores balanced across the full study.
Make the best decision based only on the evidence in that pair (audio + transcript).
    \end{lstlisting}
    \caption{The instructions given to the participants in the pairwise comparison human evaluation.}
    \label{fig:pairwisecomparison_instructions}
\end{figure*}

% analysis: Our approach is better than Moshi's approach. The reason is that Moshi makes the text too fragmented, breaking the semantic information. Our implementation of Moshi's approach and more detailed analysis is presented in Appendix xxx.

\end{document}